\documentclass[letterpaper, 10 pt, conference]{ieeeconf}  % Comment this line out if you need a4paper

\IEEEoverridecommandlockouts                              % This command is only needed if 
\usepackage[utf8]{inputenc}
\usepackage{graphics} % for pdf, bitmapped graphics files
\usepackage{epsfig} % for postscript graphics files
\usepackage{mathptmx} % assumes new font selection scheme installed
\usepackage{times} % assumes new font selection scheme installed
\usepackage{amsmath} % assumes amsmath package installed
\usepackage{amssymb}  % assumes amsmath package installed
\usepackage{xcolor}
\usepackage{amsmath}
\usepackage{bm}
\usepackage{graphicx}
\usepackage{subcaption}
\usepackage{tikz}
\usepackage{pifont}
\usepackage{booktabs}
\usepackage{multirow}

\usepackage{array}
\usepackage{subcaption}
\usetikzlibrary{fit,positioning,shapes.callouts}
\usepackage{float}
\usepackage{fancyhdr}
\title{\LARGE \bf
OpenTie: Open-vocabulary Sequential Rebar Tying System*}

\author{Sai Fan$^{1}$ and Mingze Liu$^{1}$ and Haozhen Li$^{1}$ \\ Haobo Liang$^{1}$ and Yixing Yuan$^{1}$ and Yanke Wang$^{1,2,*,},~\IEEEmembership{Member,~IEEE}$% <-this % stops a space
\thanks{\copyright~2026 IEEE. Personal use of this material is permitted. Permission from IEEE must be obtained for all other uses, including reprinting/republishing this material for advertising or promotional purposes, creating new collective works, resale or redistribution to servers or lists, or reuse of any copyrighted component of this work in other works.}
\thanks{This paper was funded by InnoHK-HKCRC.}% <-this % stops a space
\thanks{*Corresponding author: Yanke Wang.}
\thanks{This work was done during the internship of Sai Fan and Mingze Liu at HKCRC.}
\thanks{$^{1}$Sai Fan, Mingze Liu, Haozhen Li, Haobo Liang, and Yixing Yuan are with Hong Kong Center for Construction Robotics, The Hong Kong University of Science and Technology, Units 808 to 813 and 815, 8/F, Building 17W, Hong Kong Science Park, Pak Shek Kok, New Territories, Hong Kong SAR, China
        {\tt\small }}%
\thanks{$^{2}$Yanke Wang is now with the Division of Business and Hospitality Management (BHM), College of Professional and Continuing Education (CPCE), The Hong Kong Polytechnic University (PolyU), Hong Kong SAR, China and was with Hong Kong Center for Construction Robotics, The Hong Kong University of Science and Technology, Units 808 to 813 and 815, 8/F, Building 17W, Hong Kong Science Park, Pak Shek Kok, New Territories, Hong Kong SAR, China {\tt\small yanke.wang@cpce-polyu.edu.hk}}
}

\begin{document}

\maketitle
\thispagestyle{fancy}
\pagestyle{plain}

%%%%%%%%%%%%%%%%%%%%%%%%%%%%%%%%%%%%%%%%%%%%%%%%%%%%%%%%%%%%%%%%%%%%%%%%%%%%%%%%
\begin{abstract}

% This electronic document is a ÒliveÓ template. The various components of your paper [title, text, heads, etc.] are already defined on the style sheet, as illustrated by the portions given in this document.
Robotic practices on the construction site emerge as an attention-attracting manner owing to their capability of tackling complex challenges, especially in the rebar-involved scenarios. Most of existing products and research are mainly focused on the collection of large amounts of data with model training demands. To fulfill this gap, we propose OpenTie, a 3D training-free rebar tying framework utilizing a RGB-to-point-cloud generation and an open-vocabulary rebar detection on the real-world test. We implement the OpenTie via a robotic arm with a binocular camera and guarantee a high accuracy by applying the prompt-based object detection method on the image filtered by our proposed post-processing procedure for the image-to-point-cloud generation framework. Our pipeline requires no training efforts and outperforms the training-based object detection, i.e., YOLO-based method, with the verification on the real-world sequential rebar tying test. The system is flexible for horizontal and vertical rebar tying tasks and holds the potential application to the real construction site with possibility of commercialization.

\end{abstract}

%%%%%%%%%%%%%%%%%%%%%%%%%%%%%%%%%%%%%%%%%%%%%%%%%%%%%%%%%%%%%%%%%%%%%%%%%%%%%%%%
\section{INTRODUCTION}
\label{sec:intro}

% \textcolor{red}{Start from writing Section \ref{sec:sys_design} and \ref{sec:exp}, before writing, read "guidance.pdf" and "IEEEtran\_bst\_HOWTO.pdf" thoroughly.} 

% \textcolor{red}{In this section, you need to include the background, why we are proposing this problem, let the readers know when this problem exists, and how important it is, the meaning of solving it, and whether it can be applied to other systems. This section will include a lot of references, papers, try to read some recent papers. Use their bibtext and copy it to "reference.bib", and cite it by using "\cite{xxx}".}

% \textcolor{red}{You can separate the section into several paragraphs as 
% \begin{enumerate}
%     \item From construction tasks to rebar tying, why rebar tying is so important. What problems and challenges exist?
%     \item The SOTA and challenges of training-free robotic manipulation. Why training-free method is so good for rebar tying.
%     \item Focus on the unsolved problems of rebar tying and training-free method, we propose what, give a summarization of our work: what proposed, what performance it has, etc.
% \end{enumerate}}

In the realm of construction engineering, rebar tying~\cite{Dababneh01012000} stands out as a critical process that ensures the structural integrity of reinforced concrete elements. However, manual rebar tying presents significant challenges~\cite{MELENBRINK2020103312}, including high labor intensity that induces worker fatigue and increases the risk of work-related accidents. These issues are exacerbated in harsh construction environments.

To address these labor-intensive challenges, robotic manipulation has emerged as a promising avenue.
With the advent of Vision-Language-Action (VLA) and Vision-Language Models (VLM), training-free robotic manipulation, often encompassing zero-shot or few-shot learning paradigms, leverages pre-trained models to enable robots to perform tasks without extensive task-specific data collection or retraining. Some SOTA examples, such as SuSIE~\cite{ICLR2024_8ea50bf4}, leverage image-editing diffusion models to generate intermediate subgoals, enabling zero-shot robotic manipulation across novel tasks.  VidBot~\cite{chen2025vidbotlearninggeneralizable3d} derives 3D affordances from monocular RGB human videos for zero-shot execution. Other approaches such as BC-Z~\cite{jang2021bc} and RoboBERT~\cite{wang2025robobertendtoendmultimodalrobotic} demonstrate zero-shot robotic manipulation by enabling robots to generalize to unseen tasks and novel language-driven action–object compositions without additional task-specific training. Despite these progresses, challenges persist, including limited generalization to novel objects or cluttered environments, difficulties in handling dynamic uncertainties, and computational demands for real-time decision-making~\cite{cui2021toward}. By enabling zero-shot adaptation, these approaches facilitate rapid deployment, improve efficiency in diverse project scales without the need for site-specific datasets~\cite{batra2025zero,han2024zero}.

So far, to the best of our knowledge, it is still an open question to explore a training-free pipeline for the autonomous sequential rebar tying task. Focusing on the challenges in rebar tying and training-free manipulation, such as precise tying in chaotic backgrounds without prior data, real-time robustness and high success rates, we propose a novel zero-shot robotic system for autonomous sequential rebar tying, OpenTie. We implement the OpenTie pipeline by using a robotic arm equipped with a binocular camera, achieving high accuracy through prompt-based object detection applied to images refined by our proposed post-processing procedure, which is built on an image-to-point-cloud generation framework for flexible rebar tying. Our contributions are summarized as follows:
\begin{enumerate}
    \item The proposal of an automated sequential rebar tying system (OpenTie),
    \item The design and implementation of the hardware and software systems,
    \item The application of a training-free object detection method and image-to-point-cloud framework to the pipeline, and
    \item Validations and comparisons with experiments on challenging sequential rebar tying tasks under different complex scenarios, demonstrating the effectiveness of the pipeline.
\end{enumerate}

This system achieves a success rate of around 90\% in real-world tests on unstructured setup. Furthermore, it demonstrates zero-shot generalization to new rebar diameters and layouts, addressing key gaps in current SOTA by minimizing deployment time and enhancing scalability for construction automation. 

The rest of the paper is organized as follows. Section \ref{sec:related} reviews the existing rebar tying robots, training-free manipulation, and vision models in 2D and 3D. Section \ref{sec:sys_design} introduces the proposed OpenTie framework with its hardware and software designs. Section \ref{sec:exp} details experimental setup, results, discussion with limitations. The conclusion is drawn in Section \ref{sec:conclusion}. 

\section{RELATED WORK}
\label{sec:related}

% \textcolor{red}{1. Recent work of rebar tying robots, systems, ideas, and analyze their drawbacks. }

% \textcolor{red}{2. Recent work of open-vocabulary robotic manipulation, and analyze their drawbacks.}

% \textcolor{red}{3. Some other important related work analysis, like point cloud from 2D image.}

\subsection{Rebar Tying Robots and Existing Automation Systems}

Automated rebar tying has made substantial progress, particularly through advancements in vision-based systems and robotic planning. Recent systems have employed RGB-D imaging combined with techniques such as Hough transform multi-segment fitting, active perception, deep learning-based keypoint detection, and enhanced point cloud registration methods to achieve accurate and flexible robotic tying operations~\cite{liu2025enhanced,tan2024rebar,jin2021robotic}. Additionally, collaborative multi-robot approaches have optimized workspace utilization through coordinated trajectory planning, enhancing system flexibility and operability~\cite{he2024study}. Lightweight models tailored for mobile platforms, such as YOLO-FAS and MobileNetV3SSD, further address computational constraints, enabling real-time detection and path planning~\cite{cheng2024vision,duan2025yolo}. Moreover, real-time rebar spacing inspection methods based on 3D keypoint detection have been integrated effectively with robotic systems, supporting automated quality control~\cite{deng20253d}.

Despite these significant advances, existing robotic rebar tying systems continue to face critical challenges. Most vision-based systems heavily rely on extensive training datasets, limiting their generalization capabilities in complex, dynamic construction environments. Moreover, current systems frequently require precise calibration and constrained operational conditions, reducing their flexibility and adaptability. The development of more generalized, robust, and easily deployable robotic solutions remains an important research direction to address these limitations comprehensively.

\subsection{Open-Vocabulary Robotic Manipulation}
Recent studies and reviews in open-vocabulary robotic manipulation show that VLM/VLA-based frameworks can interpret natural-language instructions and integrate visual perception, language understanding, and embodied control to improve instruction grounding, task planning, real-time decision-making, and adaptive execution in complex environments~\cite{din2025visionlanguageactionmodels,doi:10.36227/techrxiv.174741645.50994955/v1}. MOKA introduces a visual prompting method enabling robots to reason about keypoint affordances and generate task-specific motions~\cite{liu2024moka}. OpenAD expands affordance detection into 3D point clouds, achieving zero-shot capability by linking semantic affordances directly with geometric point cloud data~\cite{nguyen2023open}. AnyPart and OVGNet propose frameworks integrating open-vocabulary object detection with grasp pose estimation, allowing precise manipulation and robust performance on novel object categories~\cite{van2024open,li2024ovgnet}. Additionally, Point2Graph offers an end-to-end method for generating open-vocabulary 3D scene graphs purely from point cloud inputs, facilitating flexible robot navigation and interaction~\cite{xu2024point2graph}.

However, existing open-vocabulary robotic manipulation methods still face several drawbacks. These approaches often depend heavily on large-scale pretrained models, which may compromise robustness and responsiveness in dynamic, real-world environments. Furthermore, achieving precise and reliable performance in structured, repetitive industrial tasks remains challenging, indicating a clear need for advancements in model efficiency, generalization, and practical deployment capabilities.

\subsection{Visual Perception From 2D to 3D}
Extracting reliable 3D geometric information from 2D images is crucial for precise robotic manipulation tasks. Recent methods like Segment Anything Model (SAM)~\cite{kirillov2023segment} have substantially advanced general-purpose segmentation from single-view images, enabling more accurate object delineation and spatial reasoning. Approaches combining depth estimation and segmentation~\cite{deng20253d} have facilitated effective point cloud reconstruction from RGB-D sensors, simplifying the 3D perception pipeline. Nevertheless, existing perception frameworks often struggle in environments characterized by repetitive patterns and structural occlusions, such as rebar grids. Robust detection and accurate 3D reconstruction under such conditions remain challenging, requiring specialized methods capable of consistently segmenting fine-grained geometric features critical for manipulation.

Automated rebar diameter classification has been explored through point cloud-based machine learning methods. For instance, Kim et al.~\cite{kim2021automated} propose a data-driven approach for classifying rebar diameters. However, at intersections of small-diameter rebars, the scanning resolution is often insufficient, resulting in sparse and noisy point clouds that hinder accurate classification. Similarly, methods such as Liga-stereo~\cite{Guo_2021_ICCV} and TLS-based shape recognition~\cite{ishida2012shape} focus primarily on large objects in terms of area and volume. These approaches involve expensive, bulky equipment and lack validation on smaller, fine-grained targets such as steel bars, with limited experimental evidence to support their applicability. In contrast, our binocular camera setup provides a more compact, cost-effective, and flexible alternative for generating accurate point clouds in rebar-intensive scenes.

Existing point cloud segmentation methods present various trade-offs between deployment complexity, efficiency, and segmentation accuracy. Frangez et al.~\cite{frangez2021depth} employ multi-view camera setups with time-consuming calibration, resulting in high deployment overhead. In contrast, our binocular camera can generate point clouds within minutes with minimal setup. Trevor and Gedikli~\cite{trevor2013efficient} propose efficient segmentation techniques for ordered RGB-D data, but lack considerations for parallax priors or parallel-plane specialization. Möls and Li et al.~\cite{mols2020highly} introduce a highly parallelized method based on spherical convex hulls and region growing, prioritizing speed over boundary precision. However, their method does not support unified modeling and inlier selection for parallel planes. Similarly, Byrne and Taylor~\cite{byrne2009expansion} focus on obstacle avoidance and assume a single ground plane, which is not well-suited for tasks involving closely aligned parallel surfaces that demand robust planar segmentation.

In this paper, we address these challenges by integrating open-vocabulary, training-free vision-language-action pipelines with robust single-view 3D point cloud inference specifically tailored for sequential rebar tying tasks. In addition, we incorporate the training-free T-rex model~\cite{jiang2024t} to identify rectangular regions formed by rebar crossbars, ensuring flexible and efficient detection without the need for retraining. Our method demonstrates enhanced adaptability and reduced deployment complexity compared with traditional rebar tying robots, and provides reliable perception under challenging construction conditions.

\section{SYSTEM DESIGN}
\label{sec:sys_design}

The proposed OpenTie is aimed at a training-free, sequential rebar tying by using the necessary tools like robotic arm and sensor. In our frame, we conducted real-world experiments by using the UR5e robotic arm and AI stereo binocular camera. This section details the hardware design as well as the software framework. 

Additionally, a YOLO-based tying pipeline (YOLOTie) is also implemented as a control experimental group, based on the collection of a large amount of data for training.
% \textcolor{red}{1. Hardware design, including components (type, details), why do you use this device, put photos and description}

% \textcolor{red}{2. methods, which is important, including the pipeline, details of how you USE and APPLY the methods, definition of all parameters (thresholds, manually chosen number), and why do we use these parameters, etc. You do not need to give specific number of each parameters, which is usually mentioned and listed in Section \ref{sec:exp}. NOTE: write as much content that belongs to us (as novelties) as possible. Do not write the content of the existing papers.}

% \textcolor{red}{3. The evaluation metrics, including definitions and why we use them.}

\subsection{Hardware Design}

\begin{figure}[!t]
\centering
\includegraphics[width=0.45\textwidth]{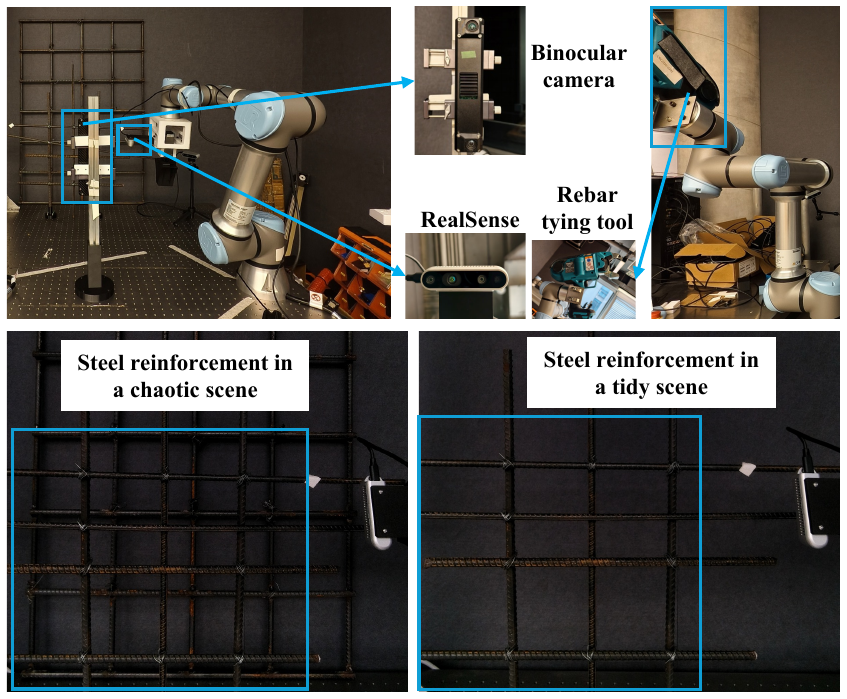}
\caption{Hardware components of OpenTie consisting of a robotic arm, a binocular camera, and a rebar tying tool. Additionally, a RealSense depth camera is used in the system of YOLOTie to conduct the comparison experiment.}
\label{fig:hardware_design}
\end{figure}

As visualized in Fig. \ref{fig:hardware_design}, the system is employed on a robotic arm, Universal Robot (UR5e), with a binocular camera set fixed out of the robot and a modified rebar tying tool installed at the end effector. The rebar tying tool, model Makita DTR181, is remolded as I/O-controlled rebar tying gun to enable automated tying functionality. The depth camera, model D435i, is used for YOLOTie and facilitates a comparative analysis of the YOLO-based object detection method against the proposed OpenTie under both chaotic and tidy scene conditions. The customized binocular camera, AI Stereo Cam, is applied for eye-to-hand calibration and point cloud generation to determine the positions of steel bars.

These hardware components jointly provide the perception capabilities and execution interfaces required for fully automatic rebar tying. We can also use this integrated solution to compare the traditional algorithm process directly based on YOLO with the proposed OpenTie framework that does not require training under different scene complexities.

% \begin{table}[h]
% \centering
% \caption{Hardware List}
% \begin{tabular}{|l|l|l|}
% \hline
% \multicolumn{1}{|c|}{\textbf{Hardware}} & \multicolumn{1}{c|}{\textbf{Type}} & \multicolumn{1}{c|}{\textbf{Num.}} \\ \hline
% Robotic arms & UR5e & 2 \\ \hline
% Rebar tying tool & Makita DTR181 & 1 \\ \hline
% Depth Camera & RealSense D435i & 1 \\ \hline
% Binocular camera & SUNWAYFOTO PC-01 & 1 \\ \hline
% \end{tabular}
% \end{table}

\subsection{Software Framework}

\begin{figure*}[h]
    \centering
    \includegraphics[width=\textwidth]{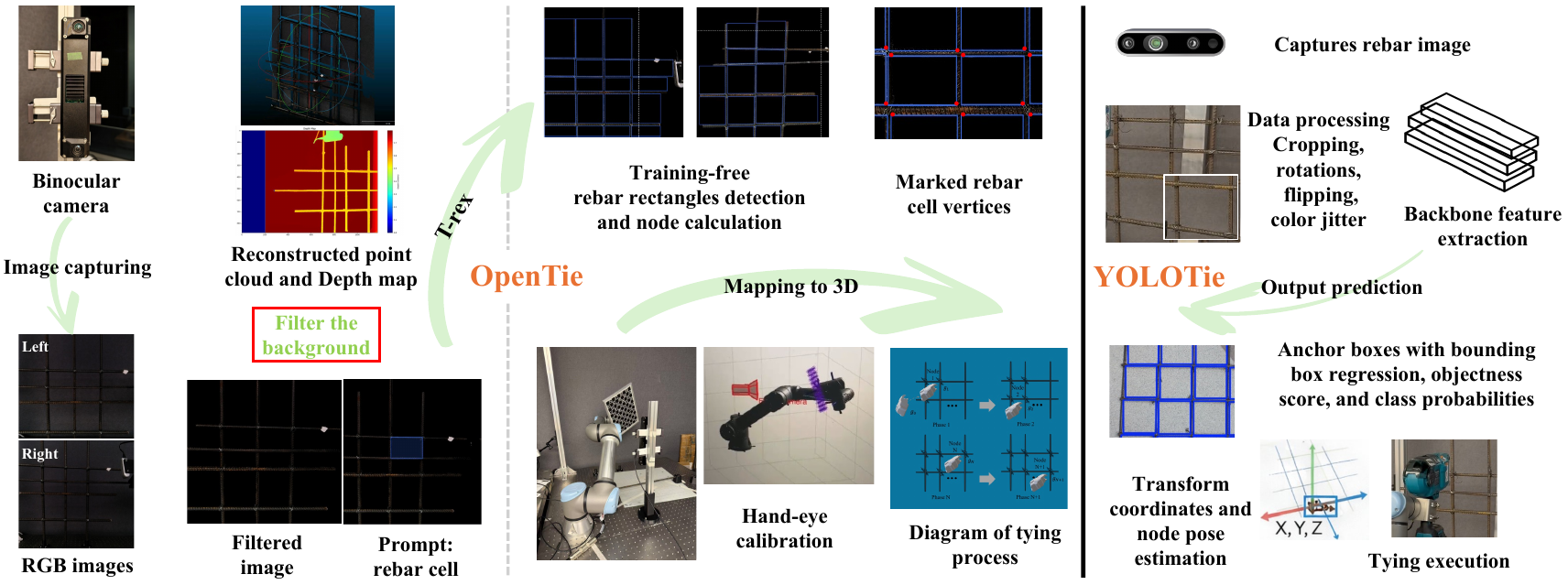} % Replace with your PDF image file

  \caption{The software diagram of the proposed OpenTie with the pipeline left, we first obtain the point cloud map by using a binocular camera and then extract the filtered rebar map. Then we obtain all the cells by using T-rex, the prompt can be any rebar cell. After that, we calculate the midpoint position of the two vertices, which is the position of the rebar node. The image on the right shows the data collection and evaluation  of YOLO-based rebar tying pipeline (YOLOTie).}
  \label{fig:software}
\end{figure*}

Two frameworks are proposed to do the sequential rebar tying in this work, i.e., YOLOTie and OpenTie, as visualized in Fig. \ref{fig:software}. In YOLOTie, YOLOv12~\cite{tian2026yolov} is utilized for rebar cell detection, followed by trajectory planning with MoveIt to reach a specified rebar node location calculated by the rebar cell for the grasping and tying task. Regarding OpenTie, a binocular camera captures two images of the rebar and reconstructs a 3D point cloud. We use a self\textcolor{green}{-} developed stereo vision camera, the specific specifications and parameters of the camera are shown in Table \ref{tab:specification}. The proposed OpenTie framework complements conventional CNN-based instruction generation methods, which have been widely used to convert visual or sensory inputs into robotic control commands~\cite{11456847},~\cite{dlugosz2023application}. While YOLO performs well in structured scenes with sufficient training data, OpenTie improves robustness in cluttered or unseen construction scenarios through training-free open-vocabulary detection and geometry-aware 3D verification.
It enables AI generation of 4K dense point cloud, based on the stereo matching algorithm. We first perform standard stereo rectification by using OpenCV and obtain the intrinsic matrix \(K\), then apply a synchronized scaling factor (\(\text{s}=0.35\)) to \(K\) to match the resized image resolution. The scaled intrinsic matrix is defined as
\begin{equation*}
    K'=sK, \quad s=0.35 .
\end{equation*}

The rectified left image is sent to a remote server to compute the disparity map. Pixels that are visible in the left view but occluded in the right view are filtered out, after which the disparity map is converted into a depth map. 
After removing occluded pixels and pixels with unreliable disparity values, the depth of each remaining pixel is computed as
\begin{equation*}
    Z(u,v)=\frac{f_x B}{d(u,v)},
\end{equation*}
where \(f_x\), \(B\), and \(d(u,v)\) denote the processed focal length, stereo baseline, and disparity value at pixel \((u,v)\), respectively. Finally, we back-project the depth map to reconstruct a 3D point cloud. The back-projection is formulated as
\begin{equation*}
    X=\frac{(u-c_x)Z}{f_x}, \quad
    Y=\frac{(v-c_y)Z}{f_y},
\end{equation*}
where \((c_x,c_y)\) is the principal point of the processed camera intrinsic matrix.

Parallel planes are then identified in the point cloud to produce a filtered image. Before plane extraction, we remove unreliable points by applying a disparity threshold \(d_{th}=90\), performing statistical outlier removal based on the 20 nearest neighbors, and conducting voxel downsampling with a voxel size of \(0.006\,\mathrm{m}\). The target rebar plane is then extracted by RANSAC~\cite{ling2024ransac} with K-Means clustering:
\begin{equation*}
    \pi:\mathbf{n}^{T}\mathbf{p}+b=0,
\end{equation*}
where \(\mathbf{n}\) is the plane normal vector, \(\mathbf{p}\) is a 3D point on the plane, and \(b\) is the plane offset.
This filtered image is automatically labeled by using T-rex to recognize the rebar cell, and vertex coordinates of the rebar cell are used to calculate the image coordinates of the binding nodes. 
Given two selected 3D vertices, the tying node is computed as
\begin{equation*}
    {}^{C}\mathbf{p}_{node}
    =
    \frac{{}^{C}\mathbf{p}_{a}+{}^{C}\mathbf{p}_{b}}{2},
\end{equation*}
where \({}^{C}\mathbf{p}_{a}\) and \({}^{C}\mathbf{p}_{b}\) are the two selected rebar-cell vertices represented in the camera coordinate system, and \({}^{C}\mathbf{p}_{node}\) is the resulting tying-node position in the camera frame.
Hand-eye calibration provides the transformation matrix to convert these binding node positions to the robotic arm's base coordinate system, incorporating a bias matrix to account for the rebar tying tool's installation position. 
The final tying position in the robot base frame is
\begin{equation*}
    {}^{B}\mathbf{p}_{tie}
    =
    {}^{B}\mathbf{T}_{C}
    \begin{bmatrix}
    {}^{C}\mathbf{p}_{node} \\
    1
    \end{bmatrix}
    +
    \Delta \mathbf{p}_{tool},
\end{equation*}
where \({}^{B}\mathbf{T}_{C}\) is the hand-eye calibration matrix from the camera frame to the robot base frame, \(\Delta \mathbf{p}_{tool}\) is the installation offset of the rebar tying tool, and \({}^{B}\mathbf{p}_{tie}\) is the executable tying position in the robot base coordinate system.

Finally, socket communication facilitates trajectory planning, enabling the robotic arm to reach the binding points accurately.

\begin{table}[t]
\centering
\caption{Specifications of the AI stereo camera used in our system.}
\label{tab:specification}
\begin{tabular}{p{3.2cm} p{4.8cm}}
\toprule
\textbf{Item} & \textbf{Value} \\
\midrule
Sensor 
& Sony IMX477, 12.3 MP, 1/2.3 \\

Baseline 
& $\mathbf{B} \approx 20~\mathrm{cm}$ \\

Original focal length 
& $f_x = 2772.66$ px, $f_y = 2771.03$ px \\

Processed focal length 
& $f_x = 970.43$ px, $f_y = 969.86$ px \\

Processing resolution 
& $1216 \times 912$ \\

Recommended range 
& $0.3$--$2.0~\mathrm{m}$ \\

Lens distortion 
& Mild barrel distortion \\

Stereo matching 
& Learning-based matching on a remote GPU server \\

Post-processing 
& Occlusion detection and depth-range filtering (without median or bilateral filtering) \\

Theoretical depth precision 
& $\approx 2.6~\mathrm{mm}$ \\

\bottomrule
\end{tabular}
\end{table}

\begin{figure*}[h]
\centering
\includegraphics[width=1.0\textwidth]{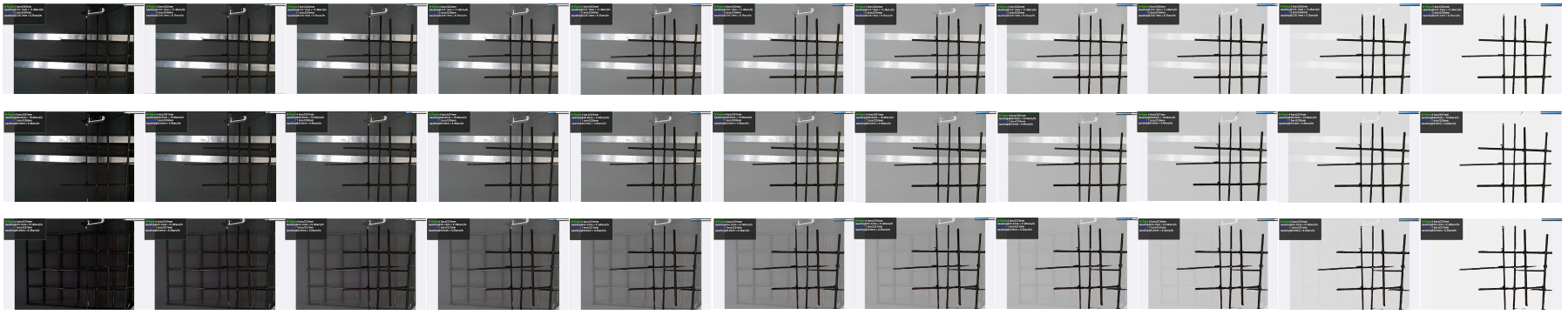}
\caption{Progressive background removal for rebar detection in three complex scenes.}
\label{fig:remove_bg}
\end{figure*}

The detailed background-filtering results are visualized in Fig. \ref{fig:remove_bg}, where the reconstructed point cloud is progressively filtered to retain the target rebar plane under complex backgrounds.

\subsection{Evaluation Metrics}

We evaluate the system via objective, repeatable measurements. For the rebar-tying node of each scene, we repeat the full perception-to-execution procedure 15 times and report localization error statistics together with a physical pull-test criterion for task success for each scene.

\textbf{Rebar Tying Success Rate (RTSR)}: 
We define a trial as successful if all tying nodes in the sequential binding process pass a pull-test under a prescribed tensile load. 
For the $i$-th trial, suppose that there are $K$ tying nodes in the scene. 
After the sequential tying operation, a constant tensile load $F_{\mathrm{test}}$ is applied to each tied node, and the corresponding displacement $\Delta d_{i,k}$ is measured for the $k$-th node, where $k = 1, 2, \dots, K$. 
The $i$-th trial is counted as successful only if all tied nodes satisfy
\[
\Delta d_{i,k} \le d_{\mathrm{th}}, \qquad k = 1, 2, \dots, K,
\]
where $d_{\mathrm{th}}$ is the displacement threshold (in mm). 
Accordingly, the success indicator for the $i$-th trial is defined as
\[
\mathrm{Success}_i = \prod_{k=1}^{K} \mathbb{I}\!\left(\Delta d_{i,k} \le d_{\mathrm{th}}\right),
\]
where $\mathbb{I}(\cdot)$ is the indicator function. 
The RTSR over $N$ repeated trials is then computed as
\[
\mathrm{RTSR} = \left(\frac{1}{N}\sum_{i=1}^{N}\mathrm{Success}_i\right)\times 100\%.
\]

\textbf{3D Localization Error (3D-LE)}: 
To quantify the accuracy of the 2D--3D mapping and coordinate transformation, we compute the 3D localization error between the predicted tying-node position and the ground-truth (GT) position in the robot base frame. 
We adopt GT-1 (robot contact-based ground truth): for each tying node $k$, the end-effector is guided to lightly touch a physical reference at the target location, and the corresponding end-effector position in the base frame is recorded as the ground-truth position $\mathbf{p}^{gt}_{s,k}$ by using robot forward kinematics. 
Let $\mathbf{p}^{pred}_{i,s,k}$ denote the predicted 3D position of node $k$ in the $i$-th trial of scene $s$. 
The Euclidean localization error (in mm) is defined as
\[
e_{i,s,k}=\left\|\mathbf{p}^{pred}_{i,s,k}-\mathbf{p}^{gt}_{s,k}\right\|_2.
\]
For each node, we report the root-mean-square error (RMSE) over $N$ repeated trials:
\[
\mathrm{3D\text{-}LE}_{s,k}
=
\sqrt{
\frac{1}{N}
\sum_{i=1}^{N}
e_{i,s,k}^{2}
}.
\]
When needed, the scene-level or overall 3D localization error is obtained by averaging the per-node RMSE values over all tying nodes in the corresponding scene or over all scenes.

\section{EXPERIMENT AND VALIDATION}
\label{sec:exp}

% \textcolor{red}{1. Why do we do these experiments. details of the experiments setup, like parameters mentioned in Section \ref{sec:sys_design}, how many experiments are done, and the computer GPU usage, inference time, etc. Two groups experiments shall be done: comparison with YOLOv12 and ablation study (deleting some parts designed by us, see the performance is reduced).}

% \textcolor{red}{2. Comparison results. tables and plots, figures, analysis, all the tables and figures shall be carefully described in detail in the texts.}

% \textcolor{red}{3. Ablation study. tables and plots, figures, analysis, all the tables and figures shall be carefully described in detail in the texts.}

% \textcolor{red}{4. Failure cases and limitations.}

% \subsection{Control Experiment}

To verify the effectiveness of the proposed OpenTie, we conducted experiments in different scenarios and made the comparison with the YOLO-based method. This section details the setup and results of the experiments.

\subsection{Experimental Setup}
\label{sec:exp_setup}
As shown in the right side of Fig. \ref{fig:software}, to experiment with the proposed OpenTie model, we utilized UR5e as the necessary tool and selected AI stereo camera as our sensor to test the success rate of binding in each scene. As for YOLOTie, we utilized an Intel RealSense camera to collect a number of rebar images for training by using the YOLO object detection algorithm. The training process was conducted in different background environments.

\begin{figure}
\centering
\includegraphics[width=0.36\textwidth]{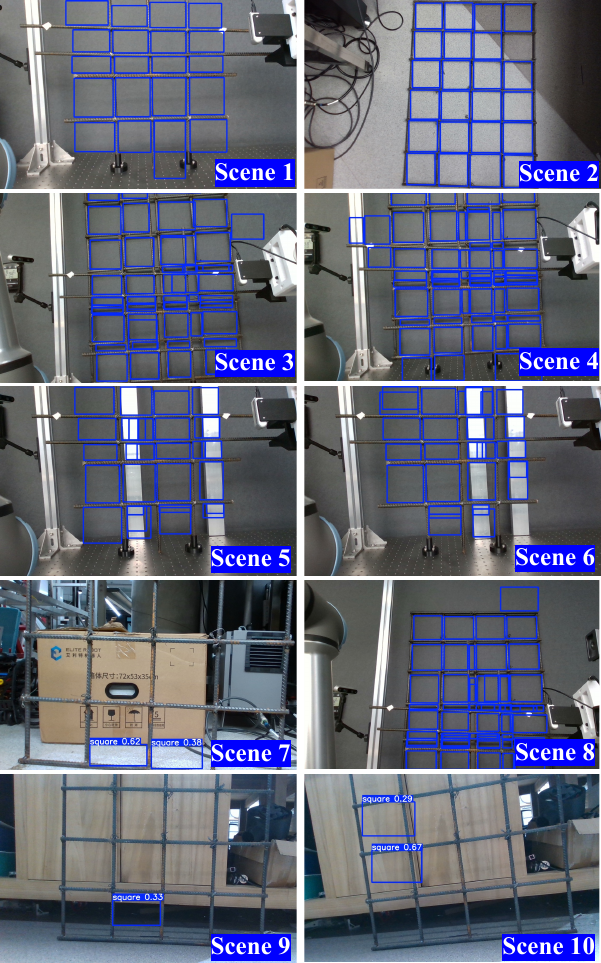}
\caption{YOLOTie detection performance across 10 scenes with varying levels of environmental complexity.
Scene 1 and 2 represent clean backgrounds (wall and floor, respectively). Scene 3 and 4 introduce additional rebar layers behind the target frame. Scenes 5, 6, 7, and 9 include significant background clutters such as metal materials, boxes, and cabinets, which hinders accuracy. Scenes 8 and 10 show slightly tilted rebar frames, leading to detection failures.}
\label{fig:bg}
\end{figure}

To examine the robustness of OpenTie and YOLOTie under varying environmental conditions, as shown in Fig. \ref{fig:bg}, we systematically evaluated the accuracy of rebar node coordinates across different rebar scenes. Particularly, Scene 1 and Scene 2 represent clean and uncluttered backgrounds, and Scenes 3 to 10 are composed of more complex and cluttered environments. In each scene, we conducted 15 repeated experiments of steel bar binding. In the simple scenes such as Scene 1 and 2, we choose to bind two nodes sequentially for testing, and in the chaotic scenario, i.e., Scene 3-10, we sequentially bind three nodes for testing. we then used the 3D-LE and RTSR indicators to conduct positioning error analysis and success rate. In the experiments, the applied constant tensile force $F_{\text{test}}$ is set to $30\,\mathrm{N}$, the displacement threshold of the tied joint $\Delta d_i$ is set to $2\,\mathrm{mm}$, and repeated trials $N$ is set to 15. After the $i$-th tying operation in each scene, a pull-test is performed to measure the displacement of the tied node under the applied load. All the tied nodes pass the pull-test, which is considered a success in this trial.

% \begin{figure}[!t]
%   \centering
%   \begin{minipage}[Pipeline]{0.175\textwidth}
%     \centering
%     \includegraphics[width=\textwidth]{hand-eye.jpg} % Replace with your PDF image file
%     \label{fig:two_images}
%   \end{minipage}
%   \hfill
%   \begin{minipage}[Flow chart]{0.275\textwidth}
%     \centering
%     \includegraphics[width=\textwidth]{plots/matlab.png}
%     \label{fig:two_images}
%   \end{minipage}
%   \caption{Left: Hand-Eye calibration setup by using UR5e. Right: Visualization of the transformation matrix result in MATLAB.}
%   \label{fig:hand_eye}
% \end{figure}

\begin{figure*}[h]
\centering
\includegraphics[width=0.7\textwidth]{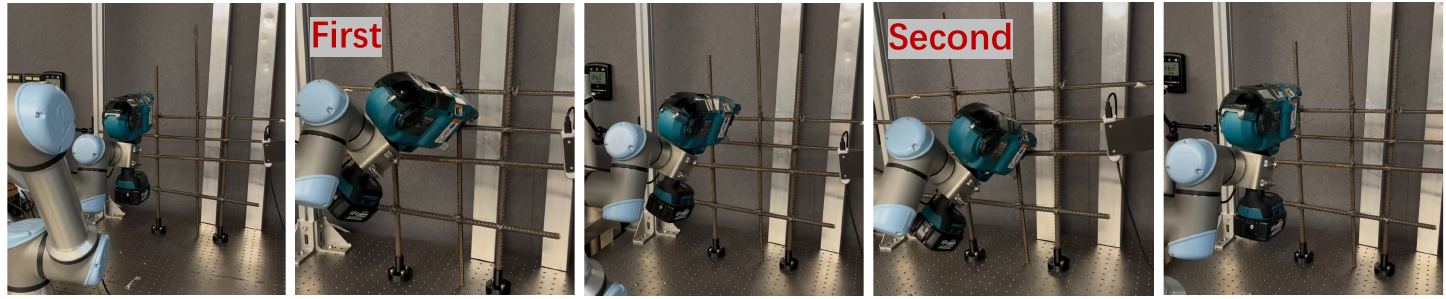}
\caption{Manipulation in simple scene, including the tying of two nodes.}
\label{fig:manipulation2}
\end{figure*}

\begin{figure*}[h]
\centering
\includegraphics[width=0.9\textwidth]{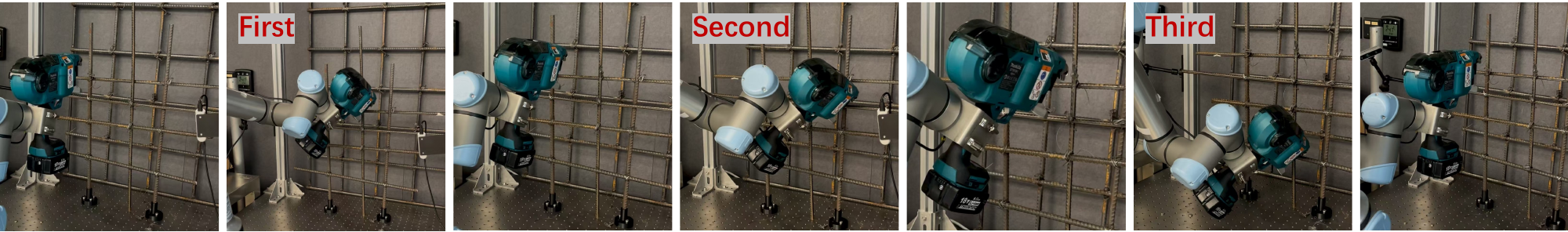}
\caption{Manipulation in chaotic scene, including the tying of three nodes.}
\label{fig:manipulation1}
\end{figure*}

In order to achieve the experimental pipeline, one important step cannot be ignored, i.e., the hand-eye calibration, as shown in Fig. \ref{fig:software}. We applied the proposed OpenTie and YOLOTie pipelines to the Universal Robots (UR5e) collaborative Robot arm. Only we have the coordinates of the rebar node obtained after calibration, our entire identification and binding process can be a completed processing. The hand-eye calibration was done by MATLAB, and we
obtained the transformation from the camera coordinate system to the robot base coordinate system to calculate the coordinates of the rebar node in the robot base coordinate system. In detail, we aim to solve for the rigid transformation $\mathbf{X}$ between the end-effector and the camera. The classic AX=XB formulation is expressed as,

\[
\underbrace{{}^{\text{End}}_{\,\text{Base1}}\mathbf{T}^{-1}
\cdot
{}^{\text{End}}_{\,\text{Base2}}\mathbf{T}}_{\mathbf{A}}
\cdot \mathbf{X}
=
\mathbf{X} \cdot
\underbrace{{}^{\text{Camera2}}_{\,\text{Object}}\mathbf{T}
\cdot
{}^{\text{Camera1}}_{\,\text{Object}}\mathbf{T}^{-1}}_{\mathbf{B}}, \text{ where}
\]
\begin{itemize}
    \item ${}^{\text{End}}_{\,\text{Base}}\mathbf{T}$: Transformation matrix of end-effector relative to robot base.
    \item ${}^{\text{Camera}}_{\,\text{Object}}\mathbf{T}$: Transformation matrix of calibration object in camera frame.
    \item $\mathbf{X}$: Fixed unknown transformation from camera to end-effector.
\end{itemize}

\noindent In practice, multiple motion pairs $(\mathbf{A}_i, \mathbf{B}_i)$ are collected and used to solve for $\mathbf{X}$ by using least square method or quaternion-based methods.

Finally, in each scenario, we conducted complete rebar tying experiments by using OpenTie and YOLOTie respectively, and recorded the success rate as well as the measurement error at the same time.

\subsection{Results}
\label{sec:exp_res}

For each node of scene, we conducted repeated tests of 15 groups of experiments. We calculated the node coordinates in the camera coordinate system and manually measured the ground-truth values of these coordinates in the camera coordinate system. By using these two values, we obtain the 3D-LE of each node in every replicate experiment, and then calculate the mean value of each node 3D-LE over 15 experiments, the results of two methods are summarized in Table. \ref{tab:node_comparison}. The results show that YOLO has a relatively locating error in node recognition under simple backgrounds, but a high error under complex backgrounds. The results indicate that in simple scenes, the results detected by YOLO tend to stabilize, but in complex scenes, the results detected by YOLO fluctuate greatly. There are also failure cases where rebar cells cannot be identified such as in scene 7, 9, 10.
% The success rate of identifying rebar cell will not increase, unless we collect more data from various scenarios and conduct extensive training. Compared with the OpenTie we proposed, it undoubtedly adds a lot of time and labor costs.

% \begin{figure}
% \centering
% \includegraphics[width=0.45\textwidth]{plots/figure1.pdf}
% \caption{Average detection accuracy across 10 scenes by using YOLOTie. Each scene includes 15 trials, the average accuracy is computed as the mean of per-trial detection accuracies, and the results show that accuracy decreases as scene complexity increases.}
% \label{fig:flow chart1}
% \end{figure}

% \begin{figure}
% \centering
% \includegraphics[width=0.45\textwidth]{plots/figure2.pdf}
% \caption {Per-trial detection accuracy in Scenes 1–10 by using YOLOTie. Each scene consists of 15 independent trials.}
% \label{fig:flow chart2}
% \end{figure}

 Because YOLO requires a large amount of data collection and training, we chose to use a training-free method T-rex to identify the rebar cell. Finally, we controled the wire gun automatically to achieve the tying of the reinforcing bars. We switched to different backgrounds to tie the rebars. The average success rate of binding was over 90\%, and Fig. \ref{fig:manipulation2}-\ref{fig:manipulation1} vividly list the sequential key frames for the tying of two nodes in the simple-background scene and three nodes in the chaotic-background scene, with multiple vertically-positioned nodes involved. The failures of OpenTie are more likely due to the higher sensitivity of the pull-test to parameter setting and the physical execution uncertainty in complex scenes. The superficially acceptable ties may still fail because of slightly inadequate tightening, or uneven local force distribution, indicating a limited mechanical robustness margin rather than a perception error.

In general, it is necessary to collect a large amount of data in every scenario for annotation and training when we want YOLOTie to achieve better results. However, by using the proposed OpenTie method, we only need to perform point cloud reconstruction and filtering processing for each scene once. We can obtain the position of each rebar node training-free through T-rex, and the success rate is over 90\%.

\begin{figure}[h]
\centering
\includegraphics[width=0.45\textwidth]{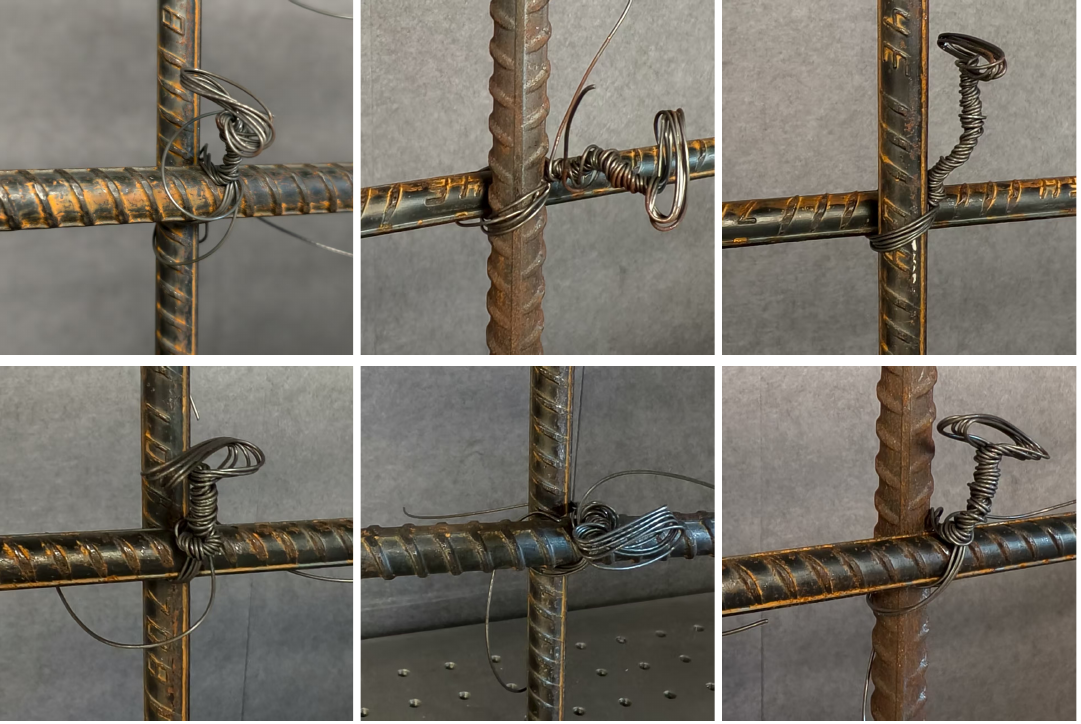}
\caption{The visualization of the tied nodes by our proposed OpenTie pipeline}
\label{fig:nodes}
\end{figure}

This experimental visualization is a proof that the proposed OpenTie works simply and effectively in the continuous rebar-tying task. Additionally, we list the images of the tied nodes in Fig. \ref{fig:nodes} for an extra proof of the valid automatic tying process after pull-test, which confirms the validity of the proposed OpenTie.

\begin{table*}[t]
\centering
\caption{Comparison between YOLOTie and OpenTie in terms of scene-level \textbf{RTSR} (\%) and node-level \textbf{3D-LE} (mm). RTSR is computed over 15 repeated trials for each scene, and 3D-LE is reported as the mean error of each tying node over the 15 repeated trials.}
\label{tab:node_comparison}

\vspace{0.3em}
\resizebox{\textwidth}{!}{%
\begin{tabular}{l*{4}{c cc c}}
\toprule
\multirow{2}{*}{\textbf{Method}}
& \multicolumn{3}{c}{\textbf{S1}}
& \multicolumn{3}{c}{\textbf{S2}}
& \multicolumn{4}{c}{\textbf{S3}}
& \multicolumn{4}{c}{\textbf{S4}} \\
\cmidrule(lr){2-4}
\cmidrule(lr){5-7}
\cmidrule(lr){8-11}
\cmidrule(lr){12-15}
& RTSR & N1 & N2
& RTSR & N1 & N2
& RTSR & N1 & N2 & N3
& RTSR & N1 & N2 & N3 \\
\midrule
YOLOTie
& 93.3 & 1.62 & 1.75
& 93.3 & 1.84 & 1.96
& 60.0 & 2.31 & 2.58 & 2.79
& 53.3 & 2.66 & 2.91 & 3.14 \\
OpenTie
& \textbf{100.0} & 1.41 & 1.48
& \textbf{100.0} & 1.52 & 1.57
& \textbf{93.3} & 1.61 & 1.68 & 1.75
& \textbf{93.3} & 1.72 & 1.78 & 1.85 \\
\bottomrule
\end{tabular}%
}

\vspace{1.0em}

\resizebox{\textwidth}{!}{%
\begin{tabular}{l*{3}{c ccc}}
\toprule
\multirow{2}{*}{\textbf{Method}}
& \multicolumn{4}{c}{\textbf{S5}}
& \multicolumn{4}{c}{\textbf{S6}}
& \multicolumn{4}{c}{\textbf{S7}} \\
\cmidrule(lr){2-5}
\cmidrule(lr){6-9}
\cmidrule(lr){10-13}
& RTSR & N1 & N2 & N3
& RTSR & N1 & N2 & N3
& RTSR & N1 & N2 & N3 \\
\midrule
YOLOTie
& 46.7 & 3.08 & 3.36 & 3.58
& 40.0 & 3.41 & 3.73 & 3.95
& 0.0 & -- & -- & -- \\
OpenTie
& \textbf{93.3} & 1.81 & 1.90 & 1.98
& \textbf{93.3} & 1.93 & 2.02 & 2.09
& \textbf{93.3} & 2.01 & 2.12 & 2.21 \\
\bottomrule
\end{tabular}%
}

\vspace{1.0em}

\resizebox{\textwidth}{!}{%
\begin{tabular}{l*{3}{c ccc}cc}
\toprule
\multirow{2}{*}{\textbf{Method}}
& \multicolumn{4}{c}{\textbf{S8}}
& \multicolumn{4}{c}{\textbf{S9}}
& \multicolumn{4}{c}{\textbf{S10}}
& \multicolumn{2}{c}{\textbf{Overall Avg.}} \\
\cmidrule(lr){2-5}
\cmidrule(lr){6-9}
\cmidrule(lr){10-13}
\cmidrule(lr){14-15}
& RTSR & N1 & N2 & N3
& RTSR & N1 & N2 & N3
& RTSR & N1 & N2 & N3
& RTSR & 3D-LE \\
\midrule
YOLOTie
& 40.0 & 4.20 & 4.47 & 4.71
& 0.0 & -- & -- & --
& 0.0 & -- & -- & --
& 42.7 & 3.06 \\
OpenTie
& \textbf{86.7} & 2.10 & 2.19 & 2.28
& \textbf{86.7} & 2.22 & 2.34 & 2.45
& \textbf{86.7} & 2.35 & 2.48 & 2.60
& \textbf{92.7} & 1.95 \\
\bottomrule
\end{tabular}%
}

\end{table*}

% \begin{figure*}[!t]
%   \centering
%   \begin{subfigure}{\textwidth}
%     \centering
%     \includegraphics[width=0.9\textwidth]{plots/manipulation2.pdf}
%     \caption{Manipulation in chaotic scene, including the tying of two nodes.}
%     \label{fig:manipulation2}
%   \end{subfigure}\hfill
%   \begin{subfigure}{\textwidth}
%     \centering
%     \includegraphics[width=\textwidth]{plots/manipulation1.pdf}
%     \caption{Manipulation in simple scene, including the tying of three nodes.}
%     \label{fig:manipulation1}
%   \end{subfigure}
%   \caption{Robotic tying of rebar nodes using the OpenTie system (the system demonstrates precise and fully autonomous tying of rebar nodes, as guided by the detected positions)}
%   \label{fig:manipulation}
% \end{figure*}

% \subsection{Evaluation metrics}

\subsection{Discussion and Limitation}
\label{sec:exp_dis}
To systematically analyze failures in chaotic scenes, we divide OpenTie failures into perception-related and execution-related errors. Perception-related errors are mainly caused by incomplete point-cloud reconstruction, occlusion, or node-position shifts around thin or overlapping rebars, especially in Scenes 8-10. Execution-related errors occur after valid node detection, including slight tool-pose deviation, insufficient tightening, uneven wire contact, or pull-test failure. Among the 11 failed OpenTie trials out of 150 repeated trials, 4 were perception-related and 7 were execution-related, indicating that most failures came from the manipulation stage rather than visual detection.

OpenTie achieved 100\% RTSR in clean Scenes 1 and 2, and the most challenging cases were Scenes 8-10 due to cluttered backgrounds and tilted rebar frames. Compared with YOLOTie, which failed to detect valid rebar cells in several chaotic scenes, OpenTie maintained more stable localization through training-free open-vocabulary detection and geometry-aware point-cloud filtering. However, current experiments were conducted in a controlled laboratory environment with real-site-like complexities, including cluttered backgrounds, overlapping rebar layers, and tilted frames. Future work will validate the system on real construction sites to improve robustness under dynamic site conditions.

\section{CONCLUSIONS}
\label{sec:conclusion}
In construction sites, steel bars are usually in a rather complex environment. In response to the high demand of sequential rebar tying system for vertically-positioned rebar structure, we propose a training-free pipeline OpenTie to achieve accurate rebar detection, localization, and tying in 3D by using low-cost camera setup with image-to-point-cloud generation. We design the hardware system via a UR5e robotic arm and the software system by implementing the proposed OpenTie as well as the YOLO-based pipeline for comparison. In the case where YOLO is used, a considerable amount of manual effort is required for annotation, and a certain amount of computing power is needed for training. The use of OpenTie can solve the problems of insufficient computing power and human resources. Moreover, our camera can generate point clouds of steel bars, allowing us to segment the point clouds and obtain the desired normal planes for recognition, which is conducive to the identification of steel bar nodes. The experiments prove that OpenTie can achieve better rebar detection than YOLOTie and the success ratio is over 90\%. The future step is to test the proposed system on the real site and we believe the OpenTie can change the manner of human tying rebars on site, especially for those vertically positioned.

\bibliographystyle{IEEEtran}
\bibliography{reference}
\end{document}